\documentclass[11pt,a4paper]{article}
\usepackage[hyperref]{emnlp2020}
\usepackage{times}
\usepackage{latexsym}
\usepackage{amsmath}
\usepackage{amssymb}
\usepackage{url}
\usepackage{textcomp}
\usepackage{graphicx}
\usepackage{placeins}

\usepackage{color,soul}

\aclfinalcopy % Uncomment this line for the final submission

\usepackage{microtype}

\title{Making Monolingual Sentence Embeddings Multilingual using Knowledge Distillation}

\author{Nils Reimers and Iryna Gurevych \\
Ubiquitous Knowledge Processing Lab (UKP-TUDA)\\
Department of Computer Science, Technische Universit\"at Darmstadt\\
\url{www.ukp.tu-darmstadt.de}}

\date{}

\begin{document}
\maketitle
\begin{abstract}
We present an easy and efficient method to extend existing sentence embedding models to new languages. This allows to create multilingual versions from previously monolingual models. The training is based on the idea that a translated sentence should be mapped to the same location in the vector space as the original sentence. 
We use the original (monolingual) model to generate sentence embeddings for the source language and then train a new system on translated sentences to mimic the original model. Compared to other methods for training multilingual sentence embeddings, this approach has several advantages: It is easy to extend existing models with relatively few samples to new languages, it is easier to ensure desired properties for the vector space, and the hardware requirements for training are lower. We demonstrate the effectiveness of our approach for 50+ languages from various language families. Code to extend sentence embeddings models to more than 400 languages is publicly available.\footnote{Code, models, and datasets: \url{https://github.com/UKPLab/sentence-transformers}}
\end{abstract}

\section{Introduction}

Mapping sentences or short text paragraphs to a dense vector space, such that similar sentences are close, has wide applications in NLP. However, most existing sentence embeddings models are monolingual, usually only for English, as applicable training data for other languages is scarce. For multi- and cross-lingual scenarios, only few sentence embeddings models exist.

In this publication, we present a new method that allows us to extend existing sentence embeddings models to new languages. We require a teacher model $M$ for source language $s$ and a set of parallel (translated) sentences $((s_1, t_1),...,(s_n, t_n))$ with $t_i$ the translation of $s_i$. Note, the $t_i$ can be in different languages. We train a new student model $\hat M$ such that $\hat M(s_i) \approx M(s_i)$ and $\hat M(t_i) \approx M(s_i)$ using mean squared loss. We call this approach \textbf{multilingual knowledge distillation}, as the student $\hat M$ distills the knowledge of the teacher $M$ in a multilingual setup. We demonstrate that this type of training works for various language combinations as well as for multilingual setups. We observe an especially high improvement of up to 40 accuracy points for low resource languages compared to LASER \cite{LASER}.

The student model $\hat M$ learns a multilingual sentence embedding space with two important properties: 1) Vector spaces are aligned across languages, i.e., identical sentences in different languages are close, 2) vector space properties in the original source language from the teacher model $M$ are adopted and transferred to other languages.

\begin{figure*}[t]
	\centering
	\includegraphics[width=0.8\linewidth]{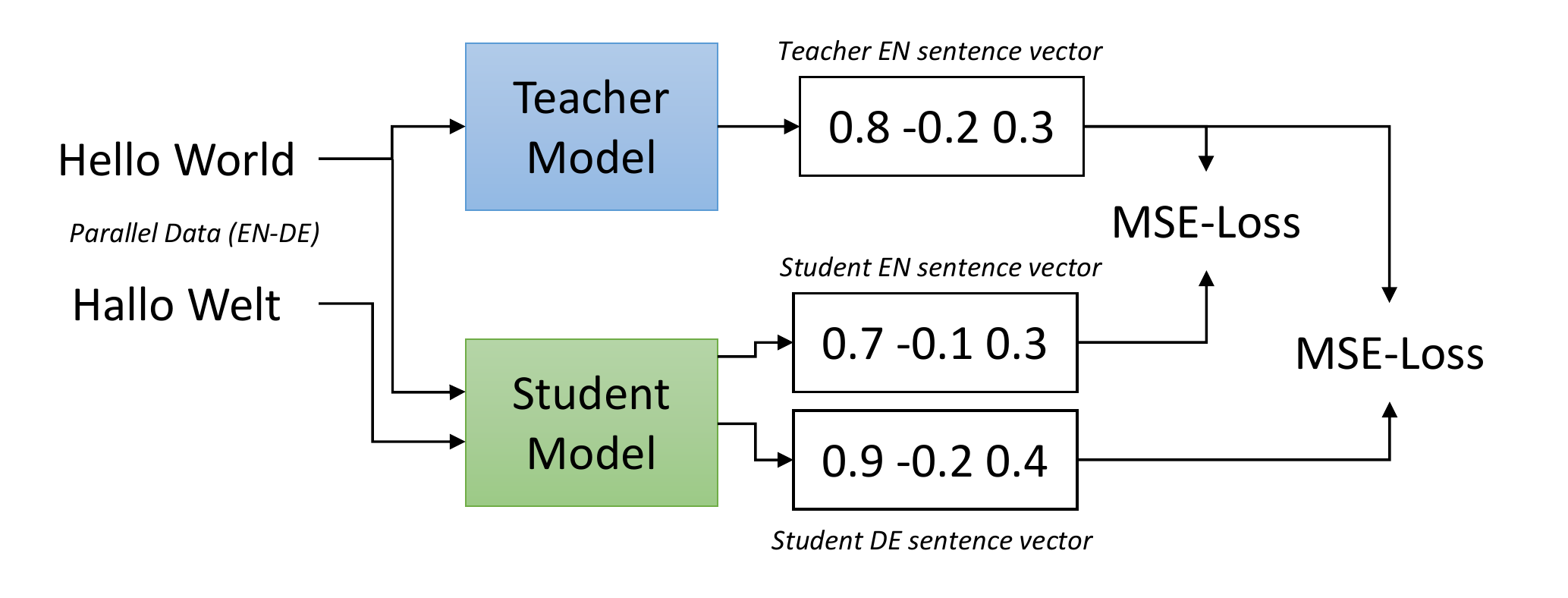}
	\caption{Given parallel data (e.g.\ English and German), train the student model such that the produced vectors for the English and German sentences are close to the teacher English sentence vector.}
	\label{fig_learning}
\end{figure*}

The presented approach has various advantages compared to other training approaches for multilingual sentence embeddings. LASER \cite{LASER} trains an encoder-decoder LSTM model using a \textbf{translation task}. The output of the encoder is used as sentence embedding. While LASER works well for identifying exact translations in different languages, it works less well for assessing the similarity of sentences that are not exact translations.  

Multilingual Universal Sentence Encoder (mUSE) \cite{mUSE1, mUSE2} was trained in a multi-task setup on SNLI \cite{snli} and on over a billion question-answer pairs from popular online forums and QA websites. In order to align the cross-lingual vector spaces, mUSE used a \textbf{translation ranking task}. Given a translation pair $(s_i, t_i)$ and various alternative (incorrect) translations, identify the correct translation. First, multi-task learning is difficult since it can suffer from catastrophic forgetting and balancing multiple tasks is not straight-forward. Further, running the \textit{translation ranking task} is complex and results in a huge computational overhead. Selecting random alternative translations usually leads to mediocre results. Instead, \textit{hard negatives} \cite{guo-etal-2018-effective} are required, i.e., alternative incorrect translations that have a high similarity to the correct translation. To get these hard negatives, mUSE was first trained with random negatives samples, then, this preliminary sentence encoder was used to identify hard negative examples. They then re-trained the network. 

In this work, we use Sentence-BERT (SBERT) \cite{sbert}, which achieves state-of-the-art performance for various sentence embeddings task. SBERT is based on transformer models like BERT \cite{devlin2018bert} and applies mean pooling on the output. In our experiments we use XLM-R \cite{xlm-r}, a pre-trained network  on 100 languages, as student model. Note, the described approach is not limited to be used with transformer models and should also work  with other network architectures.

\section{Training}
We require a teacher model $M$, that maps sentences in one or more source languages $s$ to a dense vector space. Further, we need parallel (translated) sentences $((s_1, t_1),...,(s_n, t_n))$ with $s_i$ a sentence in one of the source languages and $t_i$ a sentence in one of the target languages.

We train a student model $\hat M$ such that $\hat M(s_i) \approx M(s_i)$ and $\hat M(t_i) \approx M(s_i)$. For a given mini-batch $\mathcal{B}$, we minimize the mean-squared loss:
$$\frac{1}{|\mathcal{B}|} \sum_{j \in \mathcal{B}} \left[ (M(s_j) - \hat M(s_j))^2 +  (M(s_j) - \hat M(t_j))^2 \right]$$

$\hat M$ could have the structure and the weights of $M$, or it can be a different network architecture. This training procedure is illustrated in Figure \ref{fig_learning}. We denote trained models with $\hat M \leftarrow M$, as the student model $\hat M$ learns the representation of the teacher model $M$.

In our experiments, we mainly use an English SBERT model as teacher model $M$ and use XLM-RoBERTa (XLM-R) as student model $\hat M$. The English BERT models have a wordpiece vocabulary size of 30k mainly consisting of English tokens. Using the English SBERT model as initialization for $\hat M$ would be suboptimal, as most words in other latin-based languages would be broken down to short character sequences, and words in non-latin alphabets would be mapped to the \texttt{UNK} token. In contrast, XLM-R uses SentencePiece\footnote{\url{https://github.com/google/sentencepiece}}, which avoids language specific pre-processing. Further, it uses a vocabulary with 250k entries from 100 different languages. This makes XLM-R much more suitable  for the initialization of the multilingual student model.

\section{Training Data}
In this section, we evaluate the importance of training data for making the sentence embedding model multilingual. The OPUS website\footnote{\url{http://opus.nlpl.eu/}} \cite{TIEDEMANN12} provides parallel data for hundreds of language pairs. In our experiments, we use the following datasets:
\begin{itemize}
	\item \textbf{GlobalVoices:} A parallel corpus of news stories from the web site Global Voices.
	\item \textbf{TED2020:} We crawled the translated subtitles for about 4,000 TED talks, available in over 100 languages. This dataset is available in our repository.
	\item \textbf{NewsCommentary:} Political and economic commentary crawled from the web site Project Syndicate, provided by WMT.
	\item \textbf{WikiMatrix:} Mined parallel sentences from Wikipedia in different languages \cite{WikiMatrix}. We only used pairs with scores above 1.05, as pairs below this threshold were often of bad quality. 
	\item \textbf{Tatoeba:}  Tatoeba\footnote{\url{https://tatoeba.org/}} is a large database of example sentences and translations to support language learning. 
	\item \textbf{Europarl:} Parallel sentences extracted from the European Parliament website \cite{Europarl}.
	\item \textbf{JW300:} Mined, parallel sentences from the magazines \textit{Awake!} and \textit{Watchtower} \cite{jw300}.
	\item \textbf{OpenSubtitles2018:} Translated movie subtitles from opensubtitles.org \cite{OpenSubtitles2016}.
	\item \textbf{UNPC:} Manually translated United Nations documents from 1994 - 2014  \cite{UNPC}.
\end{itemize}

Getting parallel sentence data can be challenging for some low-resource language pairs. Hence, we also experiment with bilingual dictionaries:

\begin{itemize}
	\item \textbf{MUSE:} MUSE\footnote{\url{https://github.com/facebookresearch/MUSE}} provides 110 large-scale ground-truth bilingual dictionaries created by an internal translation tool \cite{conneau2017word}. 	
	\item \textbf{Wikititles:} We use the Wikipedia database dumps to extract the article titles from cross-language links between Wikipedia articles. For example, the page "United States" links to the German page "Vereinigte Staaten". This gives a dictionary covering a wide range of topics. 
\end{itemize}
	
The data set sizes for English-German (EN-DE) and English-Arabic (EN-AR) are depicted in Table \ref{table_train_sts2017_en_de}. For training, we balance the data set sizes by drawing for a mini batch roughly the same number of samples from each data set. Data from smaller data sets is repeated.

We trained XLM-R as our student model and used SBERT fine-tuned on English NLI and STS data\footnote{bert-base-nli-stsb-mean-tokens model from our repository} as our teacher model. We trained for a maximum of 20 epochs with batch size 64, 10,000 warm-up steps, and a learning rate of 2e-5. As development set, we measured the MSE loss on  hold-out parallel sentences.

In \cite{reimers-gurevych-2017-reporting, reimers_single_perf_score}, we showed that the random seed can have a large impact on the performances of trained models, especially for small datasets. In the following experiments, we have quite large datasets of up to several million parallel sentences and we observed rather minor differences ($\sim$ 0.3 score points) between random seeds.

\section{Experiments} \label{sec:experiments}
In this section, we conduct experiments on three tasks: Multi- and cross-lingual semantic textual similarity (STS), bitext retrieval, and cross-lingual similarity search. STS assigns a score for a pair of sentences, while bitext retrieval identifies parallel (translated) sentences from two large monolingual corpora.

Note, evaluating the capability of different strategies to align vector spaces across languages is non-trivial. The performance for cross-lingual tasks depends on the ability to map sentences across languages to one vector space (usually the vector space for English) as well as on the properties this source vector space has. Differences in performance can then be due to a better or worse alignment between the languages or due to different properties of the (source) vector space.

 We evaluate the following systems:

\textbf{SBERT-nli-stsb:} The output of the BERT-base model is combined with mean pooling to create a fixed-sized sentence representation \cite{sbert}. It was fine-tuned on the English AllNLI (SNLI \cite{snli} and Multi-NLI \cite{multinli}) dataset and on the English training set of the STS benchmark \cite{sts2017} using a siamese network structure. 

\textbf{mBERT / XLM-R mean:} Mean pooling of the outputs for the pre-trained multilingual BERT (mBERT) and XLM-R model. These models are pre-trained on multilingual data and have a multilingual vocabulary. However, no parallel data was used.

\textbf{mBERT- / XLM-R-nli-stsb:} We fine-tuned XLM-R and mBERT on the (English) AllNLI and the (English) training set of the STS benchmark. 

\textbf{LASER:} LASER \cite{LASER} uses max-pooling over the output of a stacked LSTM-encoder. The encoder was trained in an encoder-decoder setup (machine translation setup) on parallel corpora over 93 languages.

\textbf{mUSE:} Multilingual Universal Sentence Encoder \cite{mUSE1} uses a dual-encoder transformer architecture and was trained on mined question-answer pairs, SNLI data, translated SNLI data, and parallel corpora over 16 languages.

\textbf{LaBSE:} Language-agnostic BERT Sentence Embedding (LaBSE) \cite{LaBSE} was trained similar to mUSE with a dual-encoder transformer architecture based on BERT with 6 Billion translation pairs for 109 languages. 

\textbf{mBERT- / DistilmBERT- / XLM-R $\leftarrow$ SBERT-nli-stsb:} We learn mBERT, DistilmBERT, and XLM-R to imitate the output of the English SBERT-nli-stsb model.  

\textbf{XLM-R $\leftarrow$ SBERT-paraphrases:} We train XLM-R to imitate SBERT-paraphrases, a RoBERTa model trained on more than 50 Million English paraphrase pairs.

For our multi-lingual knowledge distillation experiments, we trained a single model with parallel data for 50 languages\footnote{ar, bg, ca, cs, da, de, el, es, et, fa, fi, fr, fr-ca, gl, gu, he, hi, hr, hu, hy, id, it, ja, ka, ko, ku, lt, lv, mk, mn, mr, ms, my, nb, nl, pl, pt, pt, pt-br, ro, ru, sk, sl, sq, sr, sv, th, tr, uk, ur, vi, zh-cn, zh-tw}.

\subsection{Multilingual Semantic Textual Similarity} \label{sec_sts2017}

\begin{table*}[t]
	\centering 
	\footnotesize
	\begin{tabular}{|l|c|c|c|c|}
		\hline
		\textbf{Model} &  \textbf{EN-EN} & \textbf{ES-ES} & \textbf{AR-AR}   & \textbf{Avg.} \\ \hline
		mBERT mean & 54.4 & 56.7 & 50.9 & 54.0 \\ \hline
		XLM-R mean & 50.7 & 51.8 & 25.7 & 42.7 \\ \hline
		mBERT-nli-stsb & 80.2  & 83.9 & 65.3 & 76.5 \\ \hline
		XLM-R-nli-stsb & 78.2 & 83.1 & 64.4 & 75.3 \\ \hline \hline
		
		\multicolumn{5}{|l|}{\textbf{Knowledge Distillation}} \\ \hline
		mBERT $\leftarrow$ SBERT-nli-stsb & 82.5 & 83.0 & 78.8 & 81.4  \\ \hline
		DistilmBERT $\leftarrow$ SBERT-nli-stsb & 82.1  & 84.0 & 77.7 & 81.2 \\ \hline
		XLM-R $\leftarrow$ SBERT-nli-stsb & 82.5 & 83.5 & 79.9 & 82.0 \\ \hline
		XLM-R $\leftarrow$ SBERT-paraphrases & 88.8 & 86.3 & 79.6 & \textbf{84.6} \\ \hline \hline
		
		\multicolumn{5}{|l|}{\textbf{Other Systems}} \\ \hline
		LASER & 77.6 & 79.7 &  68.9 & 75.4 \\ \hline
		mUSE & 86.4 & 86.9 & 76.4 & 83.2 \\ \hline
		LaBSE & 79.4 & 80.8 & 69.1 & 76.4 \\ \hline 
	\end{tabular}
	\caption{Spearman rank correlation $\rho$ between the cosine similarity of sentence representations and the gold labels for STS 2017 dataset. Performance is reported by convention as $\rho \times 100$.}
	\label{table_sts2017}
\end{table*}

\begin{table*}[t]
	\centering 
	\footnotesize
	\begin{tabular}{|l|c|c|c|c|c|c|c|c|c|}
		\hline
		\textbf{Model} & \textbf{EN-AR} & \textbf{EN-DE} & \textbf{EN-TR} & \textbf{EN-ES} & \textbf{EN-FR} & \textbf{EN-IT} & \textbf{EN-NL} & \textbf{Avg.} \\ \hline
		mBERT mean  & 16.7 & 33.9 & 16.0 & 21.5 & 33.0 & 34.0 & 35.6 & 27.2  \\ \hline
		XLM-R mean & 17.4 & 21.3 & 9.2 & 10.9 & 16.6 & 22.9 & 26.0 & 17.8  \\ \hline
		mBERT-nli-stsb & 30.9 & 62.2 & 23.9 & 45.4 & 57.8 & 54.3 & 54.1 & 46.9 \\ \hline
		XLM-R-nli-stsb & 44.0 & 59.5 & 42.4 & 54.7 & 63.4 & 59.4 & 66.0 & 55.6 \\ \hline  \hline
		\multicolumn{9}{|l|}{\textbf{Knowledge Distillation}} \\ \hline 
		mBERT $\leftarrow$ SBERT-nli-stsb & 77.2 & 78.9 & 73.2 & 79.2 & 78.8 & 78.9 & 77.3 & 77.6 \\ \hline
		DistilmBERT $\leftarrow$ SBERT-nli-stsb & 76.1 & 77.7 & 71.8 & 77.6 & 77.4 & 76.5 & 74.7 & 76.0  \\ \hline
		XLM-R $\leftarrow$ SBERT-nli-stsb & 77.8 & 78.9 & 74.0 & 79.7 & 78.5 & 78.9 & 77.7 & 77.9 \\ \hline
		XLM-R $\leftarrow$ SBERT-paraphrases & 82.3  & 84.0 & 80.9 & 83.1 & 84.9 & 86.3 & 84.5 & \textbf{83.7} \\ \hline \hline
		\multicolumn{9}{|l|}{\textbf{Other Systems}} \\ \hline 
		LASER &  66.5 & 64.2 & 72.0 & 57.9 & 69.1 & 70.8 & 68.5 & 67.0 \\ \hline
		mUSE & 79.3 & 82.1 & 75.5 & 79.6 & 82.6 & 84.5 & 84.1 & 81.1 \\ \hline
		LaBSE & 74.5 & 73.8 & 72.0 & 65.5 & 77.0 &  76.9 & 75.1 & 73.5  \\ \hline 
		
	\end{tabular}
	\caption{Spearman rank correlation $\rho$ between the cosine similarity of sentence representations and the gold labels for STS 2017 dataset. Performance is reported by convention as $\rho \times 100$.}
	\label{table_sts2017_cross}
\end{table*}

The goal of semantic textual similarity (STS) is to assign for a pair of sentences a score indicating their semantic similarity. For example, a score of 0 indicates \textit{not related} and 5 indicates \textit{semantically equivalent}.

The multilingual STS 2017 dataset \cite{sts2017} contains annotated pairs for EN-EN, AR-AR, ES-ES, EN-AR, EN-ES, EN-TR. We extend this dataset by translating one sentence of each pair in the EN-EN dataset to German. Further, we use Google Translate to create the datasets EN-FR, EN-IT, and EN-NL. Samples of these machine translated versions have been checked by humans fluent in that language.

For the generate sentence embeddings we compute cosine similarity and, as recommended in \cite{Reimers2016_STS}, compute the Spearman's rank correlation $\rho$ between the computed score and the gold score.

Table \ref{table_sts2017} shows the results for the monolingual setup and  Table \ref{table_sts2017_cross} the cross-lingual setup. 

As shown before \cite{sbert}, using mBERT / XLM-R without fine-tuning yields rather poor performance. Training on English NLI \& STS data (\textit{mBERT/XLM-nli-stsb}) significantly improves the performance also for the other languages. While in the monolingual setup (Table \ref{table_sts2017}) the performance is quite competitive, we observe a significant drop for the cross-lingual setup (Table \ref{table_sts2017_cross}). This indicates that the vectors spaces are not well aligned across languages.

Using our multilingual knowledge distillation approach, we observe state-of-the-art performances for mono- as well as for the cross-lingual setup, significantly outperforming other state-of-the-art models (LASER, mUSE, LaBSE).
Even though \textit{SBERT-nli-stsb} was trained on the STSbenchmark train set, we observe the best performance by \textit{SBERT-paraphrase}, which was not trained  with any STS dataset. Instead, it was trained on a large and broad paraphrase corpus, mainly derived from Wikipedia, which generalizes well to various topics.

In our experiments, XLM-R is slightly ahead of mBERT and DistilmBERT. mBERT and DistilmBERT use different language-specific tokenization tools, making those models more difficult to be used on raw text. In contrast, XLM-R uses a SentencePiece model that can be applied directly on raw text data for all languages. Hence, in the following experiments we only report results for XLM-R.

\subsection{BUCC: Bitext Retrieval}
\begin{table*}[h]
	\centering 
	\footnotesize
	\begin{tabular}{|l|c|c|c|c|c|c|c|c|c|}
		\hline
		\textbf{Model}  & \textbf{DE-EN }& \textbf{FR-EN} & \textbf{RU-EN} & \textbf{ZH-EN} &  \textbf{Avg.} \\ \hline
		mBERT mean & 44.1 & 47.2 & 38.0 & 37.4 & 41.7  \\ \hline
		XLM-R mean &  5.2 & 6.6 & 22.1 & 12.4 & 11.6 \\ \hline
		mBERT-nli-stsb & 38.9 & 39.5 & 26.4 & 30.2 & 33.7\\ \hline
		XLM-R-nli-stsb & 44.0 & 51.0 & 51.5 & 44.0 & 47.6 \\ \hline  \hline
		
		\multicolumn{6}{|l|}{\textbf{Knowledge Distillation}} \\ \hline 
		XLM-R $\leftarrow$ SBERT-nli-stsb  & 86.8 & 84.4 & 86.3 & 85.1 & 85.7 \\ 
		\hline 
		XLM-R $\leftarrow$ SBERT-paraphrase & 90.8 & 87.1 &  88.6 & 87.8  & 88.6  \\
		\hline \hline
		
		\multicolumn{6}{|l|}{\textbf{Other systems}} \\ \hline
		mUSE &  88.5 & 86.3 & 89.1 & 86.9 & 87.7 \\ \hline 
		LASER & 95.4 & 92.4 & 92.3 & 91.7 & 93.0 \\ \hline
		LaBSE & 95.9 & 92.5 & 92.4 & 93.0 & 93.5  \\ \hline
		
	\end{tabular}
	\caption{$F_1$ score on the BUCC bitext mining task.}
	\label{table_bucc}
\end{table*}

Bitext retrieval aims to identify sentence pairs that are translations in two corpora in different languages. \newcite{guo-etal-2018-effective} showed that computing the cosine similarity of all sentence embeddings and to use nearest neighbor retrieval with a threshold has certain issues.

For our experiments, we use the BUCC bitext retrieval code from LASER\footnote{\url{https://github.com/facebookresearch/LASER/}} with the scoring function from \newcite{artetxe-schwenk-2019-margin}:

\begin{gather*}
\text{score}(x,y) = \text{margin}(\text{cos}(x,y), \\ \sum_{z \in \text{NN}_k(x)} \frac{cos(x, z)}{2k} + \sum_{z \in \text{NN}_k(y)} \frac{cos(y, z)}{2k} 
\end{gather*}

with $x, y$ the two sentence embeddings and $\text{NN}_k(x)$ denoting the $k$ nearest neighbors of x in the other language\footnote{retrieved using using faiss: \url{https://github.com/facebookresearch/faiss}}. As margin function, we use $\text{margin}(a,b) = a/b$.

We use the dataset from the BUCC mining task \cite{zweigenbaum-etal-2017-overview, ZWEIGENBAUM18.12}, with the goal of extracting parallel sentences between an English corpus and four other languages: German, French, Russian, and Chinese. The corpora consist of 150K - 1.2M sentences for each language with about 2-3\% of the sentences being parallel. The data is split into training and test sets. The training set is used to find a threshold for the score function. Pairs above the threshold are returned as parallel sentences. Performance is measured using $F_1$ score.

Results are shown in Table \ref{table_bucc}. Using mean pooling directly on mBERT / XLM-R produces low scores. While training on English NLI and STS data improves the performance for XLM-R (XLM-R-nli-stsb), it reduces the performance for mBERT. It is unclear why mBERT mean and XLM-R mean produce vastly different scores and why training on NLI data improves the cross-lingual performance for XLM-R, while reducing the performance for mBERT. As before, we observe that mBERT / XLM-R do not have well aligned vector spaces and training only on English data is not sufficient.

Using our multilingual knowledge distillation method, we were able to significantly improve the performance compared to the mBERT / XLM-R model trained only on English data. 

While LASER and LaBSE only achieve mediocre results on the STS 2017 dataset, they achieve state-of-the-art performances on BUCC outperforming mUSE and our approach. LASER and LaBSE were specifically designed to identify perfect translations across languages. However, as the STS 2017 results show, these models have issues assigning meaningful similarity scores for sentence pairs that don't have identical meaning.  

In contrast, mUSE and our approach creates vector spaces such that semantically similar sentences are close. However, sentences with similar meanings must not be translations of each other. For example, in the BUCC setup, the following pair is not labeled as parallel text:
\begin{itemize}
 \item Olympischen Jugend-Sommerspiele fanden vom 16. bis 28. August 2014 in Nanjing (China) statt. (en: \textit{Summer Youth Olympic Games took place from August 16 to 28, 2014 in Nanjing (China)})
 \item China hosted the 2014 Youth Olympic Games.
\end{itemize}

Both sentences are semantically similar, hence our model and mUSE assign a high similarity score. But the pair is not a translation, as some details are missing (exact dates and location).

These results stress the point that there is no single sentence vector space universally suitable for every application. For finding translation pairs in two  corpora, LASER and LaBSE would be the best choice. However, for the task of finding semantically similar sentence pairs, our approach and mUSE would be the better choices.

We noticed that several positive pairs are missing in the BUCC dataset. We analyzed for SBERT, mUSE, and LASER 20 false positive DE-EN pairs each, i.e., we analyzed pairs with high similarities according to the embeddings method but which are not translations according to the dataset. For 57 out of 60 pairs, we would judge them as valid, high-quality translations. This issue comes from the way BUCC was constructed: It consists of a parallel part, drawn from the News Commentary dataset, and sentences drawn from Wikipedia, which are judged as non-parallel. However, it is not ensured that the sentences from Wikipedia are in fact non-parallel. The systems successfully returned parallel pairs from the Wikipedia part of the dataset. Results based on the BUCC dataset should be judged with care. It is unclear how many parallel sentences are in the Wikipedia part of the dataset and how this affects the scores.

\subsection{Tatoeba: Similarity Search}
In this section, we evaluate the strategy for lower resource languages, where it can be especially challenging to get well-aligned sentence embeddings. For evaluation, we use the Tatoeba test set-up from LASER \cite{LASER}: The dataset consists of up to 1,000 English-aligned sentence pairs for various languages. Evaluation is done by finding for all sentences the most similar sentence in the other language using cosine similarity. Accuracy is computed for both directions (English to the other language and back). 

As before, we fine-tune XLM-R with SBERT-nli-stsb as teacher model. As training data, we use JW300, which covers over 300 languages. To make our results comparable to LASER, we reduce the training data to the same amount as used by LASER. We selected four languages with rather small parallel datasets: Georgian (KA, 296k parallel sentence pairs), Swahili (SW, 173k), Tagalog (TL, 36k), and Tatar (TT, 119k). Tagalog and Tatar were not part of the 100 languages XLM-R was pre-trained for, i.e., XLM-R has no specific vocabulary and the language model was not tuned for these languages. 

As Table \ref{table_tatoeba} shows, we observe a significant accuracy improvement compared to LASER, indicating much better aligned vector spaces between English and these languages. Even though Tagalog had the smallest dataset with 36k pairs and that XLM-R was not pre-trained for this language, we achieve high accuracy scores of 86.2 and 84.0. We conclude that our strategy also works for low resource languages and can yield a significant improvement. The results for all languages in the Tatoeba test set can be found in the appendix.

\begin{table}[t]
	\centering 
	\footnotesize
	\begin{tabular}{|l|c|c|c|c|}
		\hline
		\textbf{Model} & \textbf{KA} & \textbf{SW} & \textbf{TL} & \textbf{TT}    \\ \hline
		\multicolumn{3}{|l|}{LASER} \\ \hline
		\hspace{1mm} en $\to$ xx &  39.7 & 54.4  & 52.6  & 28.0  \\
		\hspace{1mm} xx $\to$ en & 32.2 & 60.8 & 48.5 & 34.3  \\ \hline
		\multicolumn{3}{|l|}{XLM-R $\leftarrow$ SBERT-nli-stsb}     \\ \hline 	
		\hspace{1mm} en $\to$ xx & 73.1 & 85.4 & 86.2 & 54.5    \\
		\hspace{1mm} xx $\to$ en & 71.7 & 86.7 & 84.0 & 52.3 \\ \hline
	\end{tabular}
	\caption{Accuracy on the Tatoeba test set in both directions (en to target language and vice versa).}
	\label{table_tatoeba}
\end{table}

\section{Evaluation of Training Datasets}

To evaluate the suitability of the different training sets, we trained bilingual XLM-R models for EN-DE and EN-AR on the described training datasets. English and German are fairly similar languages and have a large overlap in their alphabets, while English and Arabic are dissimilar languages with distinct alphabets.  We evaluate the performance on the STS 2017 dataset. 

The results for training on the full datasets are shown in Table \ref{table_train_sts2017_en_de}. Table \ref{table_sts2017_ted_datasizes} shows the results for training only on the first $k$ sentences of the TED2020 dataset. 

First, we observe that the bilingual models are slightly better than the model trained for 10 languages (section \ref{sec_sts2017}): 2.2 points improvement for EN-DE and 1.2 points improvement for EN-AR. \newcite{xlm-r} calls this \textit{curse of multilinguality}, where adding more languages to a model can degrade the performance as the capacity of the model remains the same. 

For EN-DE we observe only minor differences between the datasets. It appears that the domain of the training data (news, subtitles, parliamentary debates, magazines) is of minor importance. Further, only little training data is necessary.

For the dissimilar languages English and Arabic, the results are less conclusive. Table \ref{table_train_sts2017_en_de} shows that more data does not necessarily lead to better results. With the Tatoeba dataset (only 27,000 parallel sentences), we achieve a score of 76.7, while with the UNPC dataset (over 8 Million sentences), we achieve only a score of 66.1. The domain of the parallel sentences is of higher importance. The results on the reduced TED2020 dataset (Table \ref{table_sts2017_ted_datasizes}) show that the score improves slower for EN-AR than for EN-DE with more data.

\begin{table}[h]
	\centering 
	\footnotesize
	\begin{tabular}{|l|c|c|c|c|c|}
		\hline
		\textbf{Dataset} &  \textbf{\#DE} &   \textbf{EN-DE} &  \textbf{\#AR} & \textbf{EN-AR} \\ \hline
		XLM-R mean & - & 21.3 & - & 17.4  \\
		XLM-R-nli-stsb & - & 59.5 & - & 44.0 \\ \hline
		MUSE Dict & 101k  & 75.8 & 27k & 68.8 \\ 
		Wikititles Dict & 545k  & 71.4  & 748k & 67.9  \\ 
		MUSE + Wikititles & 646k & 76.0 & 775k & 69.1 \\ \hline
		GlobalVoices & 37k  & 78.1 &  29k & 68.6 \\
		TED2020 & 483k  & 80.4  & 774k & 78.0 \\
		NewsCommentary & 118k & 77.7  &  7k & 57.4 \\
		WikiMatrix &  276k  & 79.4 & 385k & 75.4 \\
		Tatoeba & 303k  & 79.5 & 27k & 76.7 \\
		Europarl & 736k  & 78.7 & - & -   \\ 
		JW300 & 1,399k  & 80.0 & 382k & 74.0 \\
		UNPC & -  & - & 8M & 66.1 \\ 
		OpenSubtitles & 21M & 79.8 & 28M & 78.8 \\  \hline
		All datasets & 25M & 81.4 & 38M & 79.0 \\ \hline
	\end{tabular}
	\caption{Data set sizes for the EN-DE / EN-AR sections. Performance (Spearman rank correlation) of XLM-R $\leftarrow$ SBERT-nli-stsb on the STS 2017 dataset.}
	\label{table_train_sts2017_en_de}
\end{table}

\begin{table}[h]
	\centering 
	\footnotesize
	\begin{tabular}{|l|c|c|}
		\hline
		\textbf{Dataset size} &   \textbf{EN-DE} & \textbf{EN-AR}  \\ \hline
		XLM-R mean & 21.3 & 17.4  \\
		XLM-R-nli-stsb  & 59.5 & 44.0  \\ \hline
		1k & 71.5 & 48.4   \\
		5k & 74.5 & 59.6 \\
		10k & 77.0 & 69.5 \\
		25k & 80.0 & 70.2 \\ \hline
		Full TED2020 & 80.4 & 78.0   \\ \hline
	\end{tabular}
	\caption{Performance on STS 2017 dataset when trained with reduced TED2020 dataset sizes.}
	\label{table_sts2017_ted_datasizes}
\end{table}

\section{Target Language Training}
In this section we evaluate whether it is better to transfer an English model to a certain target language or if training from-scratch on suitable datasets in the target language yields better results. 

For this, we use the KorNLI and KorSTS datasets from \newcite{ham2020kornli}. They translated the English SNLI \cite{snli}, MultiNLI \cite{multinli}, and STSbenchmark (STSb) \cite{sts2017} datasets to Korean with an internal machine translation system. The dev and tests were post-edited by professional translators.

Ham et al.\ fine-tuned Korean RoBERTa and XLM-R on these datasets using the SBERT framework. We use the translated sentences they provide and tuned XLM-R using multilingual knowledge distillation. We use SBERT-nli-stsb as teacher model. Results are shown in Table \ref{table_sts_korean}.

\begin{table}[h]
	\centering 
	\footnotesize
	\begin{tabular}{|l|c|}
		\hline
		\textbf{Model} &   \textbf{KO-KO}  \\ \hline
		LASER & 68.44 \\
		mUSE & 76.32 \\ \hline
		\textbf{Trained on KorNLI \& KorSTS} & \\
		Korean RoBERTa-base & 80.29 \\
		Korean RoBERTa-large & 80.49 \\
		XLM-R & 79.19 \\
		XLM-R-large & 81.84 \\ \hline
		\textbf{Multiling. Knowledge Distillation} & \\
		XLM-R $\leftarrow$  SBERT-nli-stsb & 81.47 \\ 
		XLM-R-large $\leftarrow$ SBERT-large-nli-stsb &  83.00 \\ \hline
		
	\end{tabular}
	\caption{Spearman rank correlation on Korean STSbenchmark test-set \cite{ham2020kornli}.}
	\label{table_sts_korean}
\end{table}

We observe a slight improvement of using multilingual knowledge distillation over training the models directly on the translated NLI and STS data. This is great news: Training on the Korean datasets yields a model only for Korean, while with multilingual knowledge distillation, we get a model for English and Korean with aligned vector spaces. Further, we do not necessarily have a performance drop if there is only training data for the sentence embedding method in English available.

\section{Language Bias}
\newcite{lareqa} introduces the concept of \textit{language bias}: A model prefers one language or language pair over others. For example, a model would have a language bias if it maps sentences in the same language closer in vector space just because they are of the same language. Language bias can be an issue if the task involves a multilingual sentence pool: certain language pairs might get discriminated, potentially harming the overall performance for multilingual sentence pools. 
 
Figure \ref{fig_language_bias} shows the plot of the first two principle components for different multi-lingual sentence embeddings methods. In the plot, we encoded the English premise sentences from XNLI \cite{conneau-2018-xnli} with their Russian translation. The plot shows for the LaBSE model a drastic separation between the two languages, indicating that the language significantly impacts the resulting embedding vector.

The experiments in Section \ref{sec:experiments} used so far mono-lingual sentence pools, i.e., all sentences in the source / target pool were of the same language. Hence, these benchmarks are not suited to measure a potential harmful effect from language bias. In order to measure a potential negative effect from language bias, we combine all sentence pairs from the multilingual STS dataset and compute similarity scores as described in Section \ref{sec_sts2017}. Note, in Section \ref{sec_sts2017}, the models had only to score e.g.\ EN-AR sentence pairs. Now, there are 10 language combinations in one joined set. A model without language bias would achieve on this joined set a performance similar to the average of the performances over the individual subsets. However, if a model has a language bias, sentence pairs from specific language combinations will be ranked higher than others, lowering the Spearman rank correlation for the joint set.

The results are depicted in Table \ref{table_sts_joined}. We observe that LaBSE has a difference of -1.29 and LASER has a difference of -0.92. Both scores are statistically significant with confidence $p < 0.001$. LASER and LaBSE both have a language bias which decrease the performance on multilingual pools compared to mono-lingual pools. In contrast, mUSE and the proposed multilingual knowledge distillation have a minor, statistically insignificant language bias. There, the performance for the joined set only decreases by -0.19 and -0.11 compared to the evaluation on the individual sets.

In summary, mUSE and the proposed multilingual knowledge distillation approach can be used on multilingual sentence pools without a negative performance impact from language bias, while LASER and LaBSE prefer certain language combinations over other, impacting the overall result.

\begin{table*}[h]
	\centering 
	\footnotesize
	\begin{tabular}{|l|c|c|c|}
		\hline
		\textbf{Model} & \textbf{Expected Score} & \textbf{Actual Score} & \textbf{Difference}  \\ \hline
		LASER & 69.5 & 68.6 & -0.92 \\
		mUSE & 81.7 & 81.6 & -0.19 \\
		LaBSE & 74.4 & 73.1 & -1.29 \\
		XLM-R $\leftarrow$  SBERT-paraphrases & 84.0 & 83.9 & -0.11 \\ \hline
	\end{tabular}
	\caption{Spearman rank correlation for the multilingual STS dataset. Expected score is the average over the performance on the individual sets (Table \ref{table_sts2017} \& \ref{table_sts2017_cross}). Actual score is the correlation for one joined set of sentence pairs. Models without language bias would score on the joined set similar to the average over the individual sets. The difference shows the negative impact from the language bias.}
	\label{table_sts_joined}
\end{table*}

\section{Related Work}
Sentence embeddings are a well studied area with dozens of proposed methods \cite{SkipThought, conneau2017infersent, universal_sentence_encoder, yang-2018-learning}. Most of the methods have in common that they were only trained on English. Multilingual representations have attracted significant attention in recent times. Most of it focuses on cross-lingual word embeddings \cite{ruder17-cross-ling-embeddings}. A common approach is to train word embeddings for each language separately and to learn a linear transformation that maps them to a shared space based on a bilingual dictionary \cite{Artetxe2018}. This mapping can also be learned without parallel data \cite{conneau2017word, lample2017unsupervised}. 

A straightforward approach for creating cross-lingual sentence embeddings is to use a bag-of-words representation of cross-lingual word embeddings. However, \newcite{conneau-2018-xnli} showed that this approach works poorly in practical cross-lingual transfer settings. LASER \cite{LASER} uses a sequence-to-sequence encoder-decoder architecture \cite{seq2seq} based on LSTM networks. It trains on parallel corpora from neural machine translation. To create a fixed sized sentence representation, they apply max-pooling over the output of the encoder. LASER was trained for 93 languages on 16 NVIDIA V100 GPUs for about 5 days.

Multilingual Universal Sentence Encoder (mUSE)\footnote{\url{https://tfhub.dev/google/universal-sentence-encoder}} \cite{mUSE1, mUSE2} uses a dual-encoder architecture. It was trained in a multi-task setup on SNLI \cite{snli} and over 1 Billion crawled question-answer pairs from various communities. A translation ranking task was applied: Given a sentence in the source language and a set of sentences in the target languages, identify the correct translation pair. To work well, hard negative examples (similar, but incorrect translations) must be included in the ranking task. mUSE was trained for 16 languages with 30 million steps. LaBSE \cite{LaBSE} is based on a BERT architecture and used masked language model and 6 Billion translation pairs for training. It was trained similar to mUSE with a translation ranking loss, however, without any other training data.  

In this publication, we extended Sentence-BERT (SBERT) \cite{sbert}. SBERT is based on transformer models like BERT \cite{devlin2018bert} and fine-tunes those using a siamese network structure. By using the pre-trained weights from BERT, suitable sentence embeddings methods can be trained efficiently. Multilingual BERT (mBERT) was trained on 104 languages using Wikipedia, while XLM-R \cite{xlm-r} was trained on 100 languages using CommonCrawl. mBERT and XLM-R were not trained on any parallel data, hence, their vector spaces are not aligned.

\section{Conclusion}
We presented a method to make monolingual sentence embeddings multilingual with aligned vector spaces between the languages. This was achieved by using \textit{multilingual knowledge distillation}. We demonstrated that this approach successfully transfers properties from the source language vector space (in our case English) to various target languages. Models can be extended to multiple languages in the same training process.

This stepwise training approach has the advantage that an embedding model with desired properties, for example for clustering, can first be created for a high-resource language. Then, in an independent step, it can be extended to support further languages. This decoupling significantly simplifies the training procedure compared to previous approaches. Further, it minimizes the potential language bias of the resulting model.

We extensively tested the approach for various languages from different language families. We observe that LASER and LaBSE work well for retrieving exact translations, however, they work less well assessing the similarity of sentence pairs that are not exact translations. Further, we noticed that LASER and LaBSE show a language bias, preferring some language combinations over other.

\section*{Acknowledgments}
This work has been supported by the German Research Foundation through the German-Israeli Project Cooperation (DIP, grant DA 1600/1-1 and grant GU 798/17-1) and has been funded by the German Federal Ministry of Education and Research and the Hessen State Ministry for Higher Education, Research and the Arts within their joint support of the National Research Center for Applied Cybersecurity ATHENE.

\newpage

\bibliography{emnlp-ijcnlp-2019}
\bibliographystyle{acl_natbib}

\newpage
\appendix
\section*{Appendix}

\begin{figure*}[h]
	\centering
	\includegraphics[width=0.9\linewidth]{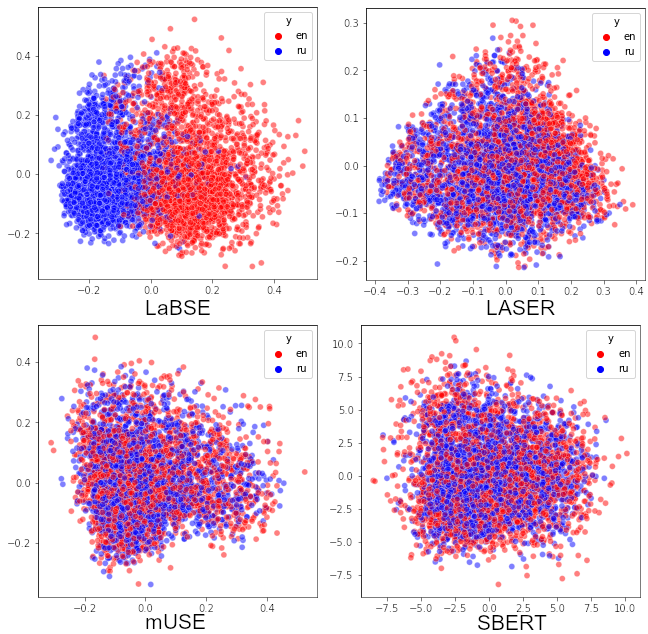}
	\caption{First two principle components for parallel sentences in English and Russian. For LaBSE, we observe a strong separation between the vector spaces, i.e., it is more biased for same language pairs. mUSE and the proposed multilingual knowledge distillation approach show nearly no language bias.}
	\label{fig_language_bias}
\end{figure*}

\section{Tatoeba Similarity Search}
LASER \cite{LASER} introduces the Tatoeba test set: It consists of up to 1,000 English-aligned sentence pairs for various languages. Evaluation is done by finding for all sentences the most similar sentence in the other language using cosine similarity. Accuracy is computed for both directions (English to the other language and back). As the two scores are usually quite close, we report the average of the scores here.

For our multi-lingual knowledge distillation, we exclude all training data from Tatoeba. This is in contrast to LASER, which used Tatoeba also for training. As the overlap between the Tatoeba test and train set is enormous, this bears the risk that the scores for LASER are artificially high. For mUSE and LaBSE, it is unknown whether they used any parallel data crawled from the Tatoeba website. 

Table \ref{table_tatoeba_all_with_data} shows the results for the languages we had parallel training data. We achieve accuracy scores usually in the 90th. Note mUSE is only available for 16 languages, hence, we also tested multilingual knowledge distillation using DistilmBERT as student and mUSE as teacher. We report this as DistilmBERT  $\leftarrow$ mUSE in the table.

For the languages which we did not use any parallel data (Table \label{table_tatoeba_all_without_data}), the scores show a much larger variance. But overall, the scores for languages without parallel data are in most cases low. This is expected, as the model did not learn how to align for these languages. Note, LaBSE and LASER had seen training data for most of the languages in Table \label{table_tatoeba_all_without_data}.
 
\begin{table*}[h]
	\centering 
	\footnotesize
	\begin{tabular}{|l|c|c|c|c|c|c|c|c|c|c|c|c|c|c|}
		\hline
		
		\textbf{Model} & \textbf{ara} & \textbf{bul} & \textbf{cat} & \textbf{ces} & \textbf{cmn} & \textbf{dan} & \textbf{deu} & \textbf{ell} & \textbf{est} & \textbf{fin} & \textbf{fra} & \textbf{glg} & \textbf{heb} \\ \hline
		LASER & 92.0 & 95.0 & 95.9 & 96.5 & 95.4 & 96.0 & 99.0 & 95.0 & 96.7 & 96.3 & 95.6 & 95.5 & 92.2 \\
		mUSE & 81.0 & 54.0 & 66.3 & 17.8 & 94.3 & 25.9 & 98.2 & 1.6 & 8.4 & 8.2 & 93.5 & 82.2 & 1.8 \\
		LaBSE & 91.0 & 95.7 & 96.5 & 97.5 & 96.2 & 96.4 & 99.4 & 96.6 & 97.7 & 97.0 & 96.0 & 97.2 & 93.0 \\
		XLM-R  $\leftarrow$ SBERT-p & 87.7 & 94.0 & 96.4 & 96.3 & 95.0 & 96.2 & 98.7 & 95.5 & 95.8 & 96.4 & 94.7 & 96.0 & 88.4 \\
		DistilmBERT  $\leftarrow$ mUSE & 86.8 & 93.3 & 95.8 & 94.6 & 95.3 & 94.6 & 98.6 & 93.1 & 93.7 & 91.9 & 94.0 & 95.2 & 85.0 \\ \hline
		\textbf{Model} & \textbf{hin} & \textbf{hrv} & \textbf{hun} & \textbf{hye} & \textbf{ind} & \textbf{ita} & \textbf{jpn} & \textbf{kat} & \textbf{kor} & \textbf{lit} & \textbf{lvs} & \textbf{mar} & \textbf{mkd} \\
		LASER & 94.7 & 97.2 & 96.0 & 36.1 & 94.5 & 95.3 & 90.7 & 35.9 & 88.9 & 96.2 & 95.4 & 91.5 & 94.7 \\
		mUSE & 1.2 & 23.9 & 10.2 & 1.7 & 93.3 & 94.3 & 93.8 & 2.6 & 86.0 & 10.2 & 11.1 & 1.8 & 33.1 \\
		LaBSE & 97.7 & 97.8 & 97.2 & 95.0 & 95.3 & 94.6 & 96.4 & 95.9 & 93.5 & 97.3 & 96.8 & 94.8 & 94.8 \\
		XLM-R  $\leftarrow$ SBERT-p & 96.4 & 97.0 & 94.7 & 91.3 & 94.1 & 94.9 & 94.2 & 91.4 & 90.1 & 95.8 & 96.4 & 91.0 & 92.2 \\
		DistilmBERT  $\leftarrow$ mUSE & 94.2 & 95.1 & 91.3 & 88.0 & 93.5 & 93.1 & 92.7 & 82.7 & 89.5 & 94.2 & 92.3 & 86.4 & 91.1 \\ \hline
		\textbf{Model} & \textbf{mon} & \textbf{nld} & \textbf{nob} & \textbf{pes} & \textbf{pol} & \textbf{por} & \textbf{ron} & \textbf{rus} & \textbf{slk} & \textbf{slv} & \textbf{spa} & \textbf{sqi} & \textbf{srp} \\ \hline
		LASER & 8.2 & 96.3 & 98.8 & 93.4 & 97.8 & 95.2 & 97.4 & 94.6 & 96.6 & 95.9 & 98.0 & 98.0 & 95.3 \\
		mUSE & 16.9 & 94.0 & 23.9 & 12.7 & 93.7 & 94.9 & 30.0 & 93.7 & 21.1 & 20.9 & 95.4 & 19.9 & 27.7 \\
		LaBSE & 96.6 & 97.2 & 98.9 & 96.0 & 97.8 & 95.6 & 97.8 & 95.3 & 97.3 & 96.7 & 98.4 & 97.6 & 96.2 \\
		XLM-R  $\leftarrow$ SBERT-p & 91.7 & 96.0 & 98.0 & 94.8 & 97.0 & 94.8 & 96.4 & 93.5 & 96.2 & 95.5 & 98.0 & 97.5 & 93.8 \\
		DistilmBERT  $\leftarrow$ mUSE & 90.6 & 95.8 & 95.8 & 90.0 & 95.3 & 94.5 & 94.7 & 94.4 & 95.4 & 94.8 & 95.5 & 95.6 & 93.2 \\ \hline
		\textbf{Model} & \textbf{swe} & \textbf{tha} & \textbf{tur} & \textbf{ukr} & \textbf{urd} & \textbf{vie} & \textbf{yue} & \textbf{zsm}  & \multicolumn{5}{c|}{} \\ \hline
		LASER & 96.6 & 95.4 & 97.5 & 94.5 & 81.9 & 96.8 & 90.0 & 96.4 & \multicolumn{5}{c|}{} \\
		mUSE & 18.8 & 96.0 & 94.0 & 51.0 & 6.4 & 10.4 & 84.2 & 89.1  & \multicolumn{5}{c|}{} \\
		LaBSE & 96.5 & 97.1 & 98.4 & 95.2 & 95.3 & 97.8 & 92.1 & 96.9  & \multicolumn{5}{c|}{} \\
		XLM-R  $\leftarrow$ SBERT-p & 95.7 & 96.3 & 97.2 & 94.3 & 92.2 & 97.2  & 84.4 & 95.6  & \multicolumn{5}{c|}{} \\
		DistilmBERT  $\leftarrow$ mUSE & 94.3 & 93.2 & 96.5 & 92.5 & 87.9 & 95.3 & 81.0 & 95.2  & \multicolumn{5}{c|}{} \\ \hline

	\end{tabular}
	\caption{Tatoeba test set results for languages \textbf{with parallel data} for multilingual knowledge distillation.}
	\label{table_tatoeba_all_with_data}
\end{table*}

\begin{table*}[h]
	\centering 
	\footnotesize
	\begin{tabular}{|l|c|c|c|c|c|c|c|c|c|c|c|c|c|c|}
		\hline
		\textbf{Model} & \textbf{afr} & \textbf{amh} & \textbf{ang} & \textbf{arq} & \textbf{arz} & \textbf{ast} & \textbf{awa} & \textbf{aze} & \textbf{bel} & \textbf{ben} & \textbf{ber} & \textbf
		{bos} & \textbf{bre} \\ \hline
		LASER & 89.5 & 42.0 & 37.7 & 39.5 & 68.9 & 86.2 & 36.1 & 66.0 & 66.1 & 89.6 & 68.2 & 96.5 & 15.8 \\
		mUSE & 63.5 & 2.1 & 38.1 & 28.2 & 59.6 & 81.5 & 2.4 & 42.2 & 40.3 & 0.7 & 8.3 & 30.1 & 10.2 \\
		LaBSE & 97.4 & 94.0 & 64.2 & 46.2 & 78.4 & 90.6 & 73.2 & 96.1 & 96.2 & 91.3 & 10.4 & 96.2 & 17.3 \\
		XLM-R  $\leftarrow$ SBERT-p & 84.5 & 67.9 & 25.0 & 30.6 & 63.7 & 78.3 & 46.5 & 85.0 & 86.9 & 77.6 & 6.8 & 95.8 & 10.1 \\
		DistilmBERT  $\leftarrow$ mUSE & 68.3 & 2.7 & 37.7 & 32.7 & 61.5 & 85.0 & 43.9 & 43.4 & 49.1 & 1.2 & 8.1 & 94.2 & 11.5 \\ \hline
		\textbf{Model} & \textbf{cbk} & \textbf{ceb} & \textbf{cha} & \textbf{cor} & \textbf{csb} & \textbf{cym} & \textbf{dsb} & \textbf{dtp} & \textbf{epo} & \textbf{eus} & \textbf{fao} & \textbf
		{fry} & \textbf{gla} \\ \hline
		LASER & 77.0 & 15.7 & 29.2 & 7.5 & 43.3 & 8.6 & 48.0 & 7.2 & 97.2 & 94.6 & 71.6 & 51.7 & 3.7 \\
		mUSE & 76.1 & 13.7 & 33.6 & 6.4 & 37.4 & 13.1 & 35.1 & 8.4 & 36.8 & 19.4 & 18.7 & 52.3 & 6.9 \\
		LaBSE & 82.5 & 70.9 & 39.8 & 12.8 & 56.1 & 93.6 & 69.3 & 13.3 & 98.4 & 95.8 & 90.6 & 89.9 & 88.8 \\
		XLM-R  $\leftarrow$ SBERT-p & 69.4 & 11.7 & 25.9 & 5.1 & 40.5 & 34.9 & 51.4 & 7.3 & 68.8 & 48.6 & 50.8 & 58.4 & 7.5 \\
		DistilmBERT  $\leftarrow$ mUSE & 77.2 & 13.8 & 34.7 & 7.3 & 48.0 & 13.1 & 52.0 & 9.4 & 41.2 & 19.0 & 36.1 & 54.0 & 6.0 \\ \hline
		\textbf{Model} & \textbf{gle} & \textbf{gsw} & \textbf{hsb} & \textbf{ido} & \textbf{ile} & \textbf{ina} & \textbf{isl} & \textbf{jav} & \textbf{kab} & \textbf{kaz} & \textbf{khm} & \textbf
		{kur} & \textbf{kzj} \\ \hline
		LASER & 5.2 & 44.4 & 54.5 & 83.7 & 86.2 & 95.2 & 95.6 & 22.9 & 58.1 & 18.6 & 20.6 & 17.2 & 7.2 \\
		mUSE & 7.7 & 39.3 & 33.3 & 55.5 & 73.3 & 86.7 & 10.3 & 38.3 & 3.7 & 15.3 & 1.5 & 21.7 & 10.2 \\
		LaBSE & 95.0 & 52.1 & 71.2 & 90.9 & 87.1 & 95.8 & 96.2 & 84.4 & 6.2 & 90.5 & 83.2 & 87.1 & 14.2 \\
		XLM-R  $\leftarrow$ SBERT-p & 18.6 & 36.8 & 57.6 & 56.0 & 70.5 & 87.9 & 75.8 & 37.3 & 2.7 & 73.7 & 64.8 & 43.7 & 8.0 \\
		DistilmBERT  $\leftarrow$ mUSE & 8.0 & 38.9 & 56.4 & 61.1 & 77.8 & 90.4 & 16.0 & 31.7 & 3.7 & 16.8 & 1.2 & 27.7 & 10.8 \\ \hline
		\textbf{Model} & \textbf{lat} & \textbf{lfn} & \textbf{mal} & \textbf{max} & \textbf{mhr} & \textbf{nds} & \textbf{nno} & \textbf{nov} & \textbf{oci} & \textbf{orv} & \textbf{pam} & \textbf
		{pms} & \textbf{swg} \\ \hline
		LASER & 58.5 & 64.5 & 96.9 & 50.9 & 10.4 & 82.9 & 88.3 & 66.0 & 61.2 & 28.1 & 6.0 & 49.6 & 46.0 \\
		mUSE & 36.7 & 60.5 & 1.2 & 65.0 & 14.3 & 57.5 & 21.2 & 66.1 & 42.9 & 28.3 & 8.4 & 48.8 & 48.7 \\
		LaBSE & 82.0 & 71.2 & 98.9 & 71.1 & 19.2 & 81.2 & 95.9 & 78.2 & 69.9 & 46.8 & 13.6 & 67.0 & 65.2 \\
		XLM-R  $\leftarrow$ SBERT-p & 28.0 & 57.7 & 94.0 & 58.5 & 11.9 & 50.7 & 89.3 & 58.8 & 52.4 & 33.4 & 7.0 & 44.3 & 33.9 \\
		DistilmBERT  $\leftarrow$ mUSE & 42.9 & 65.3 & 1.0 & 60.9 & 14.2 & 59.5 & 78.8 & 69.8 & 54.7 & 27.8 & 9.6 & 52.6 & 51.8 \\ \hline
		\textbf{Model} & \textbf{swh} & \textbf{tam} & \textbf{tat} & \textbf{tel} & \textbf{tgl} & \textbf{tuk} & \textbf{tzl} & \textbf{uig} & \textbf{uzb} & \textbf{war} & \textbf{wuu} & \textbf
		{xho} & \textbf{yid} \\ \hline
		LASER & 57.6 & 69.4 & 31.1 & 79.7 & 50.6 & 20.7 & 44.7 & 45.2 & 18.7 & 13.6 & 87.7 & 8.5 & 5.7 \\
		mUSE & 13.7 & 2.8 & 15.7 & 2.4 & 16.2 & 20.9 & 46.6 & 4.0 & 15.9 & 15.6 & 82.2 & 14.8 & 1.9 \\
		LaBSE & 88.6 & 90.7 & 87.9 & 98.3 & 97.4 & 80.0 & 63.0 & 93.7 & 86.8 & 65.3 & 90.3 & 91.9 & 91.0 \\
		XLM-R  $\leftarrow$ SBERT-p & 27.6 & 85.7 & 17.8 & 89.1 & 32.4 & 24.1 & 41.3 & 65.5 & 32.6 & 11.4 & 82.7 & 11.6 & 52.7 \\
		DistilmBERT  $\leftarrow$ mUSE & 13.8 & 2.3 & 13.5 & 1.9 & 15.7 & 27.6 & 45.2 & 4.5 & 18.7 & 15.4 & 82.2 & 13.4 & 6.2 \\ \hline

	\end{tabular}
\caption{Tatoeba test set results for languages for languages \textbf{without parallel data} for multilingual knowledge distillation. LaBSE and LASER had training data for most of these languages.}
\label{table_tatoeba_all_without_data}
\end{table*}

\end{document}